\documentclass[preprint,12pt, a4paper]{elsarticle}
\usepackage{graphicx} 

\usepackage{amsmath}
\usepackage{amssymb}
\usepackage{amsfonts}

\usepackage[T1]{fontenc}

\usepackage{url}

\usepackage[ruled,vlined]{algorithm2e}
\usepackage{algpseudocode}
\usepackage{makecell}
\usepackage{tabu}
\usepackage{todonotes}
\usepackage{booktabs}
\usepackage{multirow} 
\usepackage[export]{adjustbox}

\usepackage{comment}

\title{Computer-Aided Design of Personalized Occlusal Positioning Splints Using Multimodal 3D Data}
\author{Agnieszka Anna Tomaka} 
\author{Leszek Luchowski}
\author{\\Micha{\l} Tarnawski}
\author{Dariusz Pojda\corref{dp}}
\address{Institute of Theoretical and Applied Informatics, Polish Academy of Sciences, Ba{\l}tycka~5, 44-100~Gliwice, Poland}

\cortext[dp]{Corresponding author: dpojda@iitis.pl}

\journal{Computer Vision and Image Understanding}

\begin{document}

\begin{frontmatter}

\begin{abstract}
Digital technology plays a crucial role in designing customized medical devices, such as occlusal splints, commonly used in the management of disorders of the stomatognathic system. This methodological proof-of-concept study presents a computer-aided approach for designing and evaluating occlusal positioning splints. The primary aim is to demonstrate the feasibility and geometric accuracy of the proposed method at the preclinical stage. 
In this approach, a three-dimensional splint is generated using a transformation matrix to represent the therapeutic mandibular position. An experienced operator defines this position using a virtual patient model reconstructed from intraoral scans, CBCT, 3D facial scans, and a digitized plaster model. 
We introduce a novel method for generating splints that reproduces occlusal conditions in the therapeutic position and resolves surface conflicts through virtual embossing. The process for obtaining transformation matrices using dental tools and intraoral devices commonly employed in dental and laboratory workflows is described, and the geometric accuracy of both designed and printed splints is evaluated using profile and surface deviation analysis. The method supports reproducible, patient-specific splint fabrication and provides a transparent foundation for future validation studies, supporting multimodal image registration and quantification of occlusal discrepancies in research settings.
\end{abstract}

\begin{keyword}
Virtual Orthodontics \sep Occlusal Splint \sep Digital Modeling   \sep Designing Positioning Splints

\end{keyword}
\end{frontmatter}

\section{Introduction}

Contemporary occlusal splint design stands at the intersection of clinical practice and digital technology. While therapeutic jaw positioning remains the domain of the clinician, its realization increasingly depends on advanced tools for imaging, registration, and 3D modeling. Occlusal splints play a central role in managing stomatognathic dysfunction, serving as both therapeutic and diagnostic devices, as reported in the literature.

Conventional methods for fabricating occlusal splints, whether manual or digital, face significant limitations. Manual approaches lack reproducibility and depend heavily on individual technician skill, while many digital methods struggle with geometric conflicts such as surface collisions, loss of occlusal contact fidelity, or inadequate integration of multimodal data. These shortcomings often result in splints that do not precisely reproduce the clinician’s prescribed therapeutic mandibular position. Our approach addresses these issues by interpreting the splint as a physical realization of a rigid-body transformation, supported by a novel virtual embossing mechanism that resolves surface conflicts and preserves functional occlusion.

This work builds upon our earlier research on dynamic occlusal surfaces and splint-mediated mandibular stabilization \cite{Pojda2019, Tomaka2019}. Here, we extend the concept by preserving the natural morphology of the mandibular occlusal surface within the splint and by providing a reproducible workflow that enables precise mandibular repositioning relative to a fixed maxillary reference.

Digital splint design must reconcile medical requirements (what the splint must achieve) with technical implementation (how this is realized using algorithms and multimodal data). In the literature, two main strategies have emerged: one adapts general CAD/CAM techniques to splint construction, while the other develops dedicated tools for specific clinical objectives. Our method integrates both perspectives, combining clinical insight with algorithmic robustness.

We further develop methods for replicating natural tooth occlusion in modified jaw relations \cite{Pojda2019}, and techniques for multimodal data integration in a patient-centered coordinate system \cite{Tomaka2019, TomakaG2019}. This includes the fusion of CBCT, 3D facial and intraoral scans, facial scans with embedded dental models \cite{Tomaka2007}, and mandibular motion tracking \cite{TomakaG2019,Tomaka2016}.

Our approach models the splint as a negative impression of the dental arches, accounting for undercuts, 3D printing tolerances, and the minimum wall thickness required for structural stability. Generating the occlusal surface (the interface guiding mandibular contact) remains a particular challenge, as it must balance clinical intent with anatomical feasibility.

Crucially, the method integrates multimodal patient data into a unified coordinate system, allowing the extraction of transformation parameters and their translation into a digital splint model that realizes the intended repositioning.

In summary, this work consolidates and extends our previous contributions, integrating key stages of the digital workflow for occlusal splint design. The presented methodology demonstrates the potential of multimodal image integration for mandibular repositioning and provides a platform for future applications, such as image alignment, CR/MI analysis of discrepancies, and reproducible patient positioning.

Compared to our previous work, this study introduces an integrated workflow for computer-aided splint design that directly incorporates a user-defined therapeutic transformation matrix and a novel virtual embossing mechanism to resolve surface conflicts. The approach is validated at a preclinical, proof-of-concept stage and offers a transparent foundation for further validation studies and potential future translation into clinical workflows.

Finally, to guide the reader through the manuscript, Section~\ref{sec:literature} reviews the evolution of digital approaches to occlusal splint design. Section~\ref{sec:background} outlines the background and limitations of our previous work, motivating the need for an improved workflow. Section~\ref{sec:model} details the proposed method for constructing positioning splints, while Section~\ref{sec:results} presents their experimental validation and accuracy analysis. Section~\ref{sec:discussion} discusses broader implications and future directions, and Section~\ref{sec:conclusions} concludes with a summary of findings and their relevance for future clinical workflows and research use.

\section{ From Clinical Concept to Current Digital Practice}\label{sec:literature}

\begin{quote}
\textit{The goal of complete dentistry is defined as long term maintainable health of the stomatognathic system, with functional harmony in an environment of  healthy teeth, joints, periodontium, and musculature, and  esthetic result. Neuromuscular harmony depends on structural harmony between the occlusion and the TM joints} \cite{Dawson2007}.    
\end{quote}
The stomatognathic system consists of two jaws (the maxilla and mandible), each composed of bones, teeth, and soft tissues attached to a bony scaffolding, and it functions by altering the relationship between these jaws.
Simultaneous stable TM joints and mandibular position, along with stable occlusion, can then be treated as one of the primary goals of dentistry.

Disorders, defects, and dysfunctions of the stomatognathic system are related to its shape and structure, but also to its function. The treatment of these morphological defects or dysfunctions is often carried out with orthodontic appliances that are individually designed for each patient \cite{Okeson2020, Alqutaibi2015, Jagger2018, Klasser2009}. The variety of these orthodontic appliances is due to the diversity of needs. Thus, one can distinguish between various types of braces, expanders, aligners, retainers, prostheses, surgical and occlusal splints, mouthguards,  etc.

This paper focuses on the design of occlusal positioning splints, a class of orthodontic appliances.

\subsection{Role and clinical application of Occlusal Splint}

According to the literature, an occlusal (bite) splint is a removable dental appliance that covers the upper or lower teeth. It is commonly employed to manage centric relation (CR) and maximal intercuspation (MI) discrepancies \cite{Palaskar2013}, to support the management of temporomandibular disorders (TMD) \cite{Klasser2009}, to reduce the effects of bruxism, and to alleviate certain types of headaches \cite{Lavigne2011, Klasser2009}.

Reported clinical applications include stabilizing the stomatognathic system by maintaining proper jaw alignment and redistributing occlusal contacts, reducing TMD-related symptoms such as muscle tension or joint strain, and mitigating tooth wear. In sports, mouthguards and related splints are used to protect teeth and temporomandibular joints by absorbing impact forces. Splint therapy has been described in the literature as non-invasive and reversible, with improvements in symptoms often reported as diagnostic indicators to guide further prosthetic or orthodontic treatment planning \cite{Okeson2020}.

\subsection{Types of Occlusal Splints}

The literature describes several types of occlusal splints. Among the most common are stabilizing splints, reported to reduce teeth grinding and relax jaw muscles, and repositioning splints, designed to adjust jaw position in order to improve function \cite{Okeson2020}. A comprehensive overview of oral appliances, including their benefits and limitations, is provided in \cite{Klasser2009,Alqutaibi2015, Jagger2018}.

The literature also describes a technical classification of occlusal splints based on how the opposing teeth contact the splint surface \cite{Jagger2018}. Although direct tooth-to-tooth contact is prevented by the splint itself, this classification considers the pattern of occlusal contacts on the splint’s surface. Two main categories are distinguished: partial occlusal contact splints, such as anterior bite planes or sleep clench inhibitors, and full occlusal contact splints, such as the Michigan splint (maxillary arch) or the Tanner appliance (mandibular arch).

A commonly noted drawback of splints without full occlusal contact is the possible eruption of unsupported teeth, which may lead to an open bite \cite{Jagger2018}. Other reports indicate that reduced occlusal contact can affect elevator muscle forces: molar contact maintains up to 100\% of clenching force, cuspid contact around 60\%, and incisor contact only 20--30\% \cite{Wilkerson}.

Another frequently cited classification factor is the extent to which jaw relationships are altered. In this context, anterior (protrusive) splints advance the mandible, while stabilization splints maintain the current jaw relationship.

In summary, the design of an occlusal splint involves several factors: the arch for which the splint is intended (maxillary or mandibular), the portion of the arch it covers (partial vs.~full), and the intended interarch relationship. This relationship can be static, as in surgical wafers \cite{Chen2016}, restrictive to prevent dislocation of the TMJ disc, or adaptive, allowing controlled variation within a therapeutic range.

\subsection{Traditional manufacturing of occlusal splints}

Before the introduction of digital methods, occlusal splints were fabricated using manual techniques that demanded a high level of precision and experience \cite{Martin2015,Okeson2020}. Dental technicians worked with plaster models created from alginate impressions of both jaws and used wax patterns to establish occlusal relationships. Based on these models, the insertion path and undercuts were identified to ensure the splint could be correctly placed and removed. Articulators, simulating mandibular movements, and facebows, determining the jaw's position relative to the temporomandibular joint, were employed to enhance accuracy. Splints were shaped manually from acrylic materials, which were then thermally or chemically processed. Final fitting and surface adjustments were made intraorally, often requiring time-consuming clinical corrections. These traditional methods lacked standardization and depended heavily on individual technician skill, leading to what was reported as low repeatability in outcomes \cite{Lauren2008}.

\subsection{Digital manufacturing of occlusal splints}

The emergence of Computer-Aided Design/Computer-Aided Manufacturing (CAD/CAM) technologies has significantly transformed the workflow for occlusal splint design and fabrication. Digital imaging tools, such as CBCT, intraoral scanners, and facial scanners, provide detailed 3D models of the patient’s anatomy. Additionally, motion capture systems and devices for measuring occlusal forces enhance the precision of functional assessments. Based on this digital data, splints can be designed virtually and produced using either additive (3D printing) or subtractive (milling) techniques. Importantly, the development of CAD/CAM tools for splint design has paralleled the evolution of digital systems for surgical planning, as both rely on similar algorithms and data types.

\subsection{CAD/CAM-based design approaches}

Initially, any CAD software capable of manipulating 3D solids was considered sufficient for splint design. Publications often claimed CAD/CAM use without specifying the software or algorithms, assuming that experienced technicians could adapt general CAD tools for dental applications. Early research focused on integrating volumetric CT data with 3D surface scans to create composite models for surgical planning. Gateno et al. \cite{Gateno2003} demonstrated the clinical advantage of combining these data types for improved accuracy. Lin et al. \cite{Lin2006} investigated the feasibility of CAD/CAM-generated splints and reported favorable outcomes, including reduced manual work.

\subsection{Semi-automatic methods}

With time, more algorithmic and reproducible methods emerged. Authors began describing the use of specific computer graphics techniques such as triangular mesh processing, 3D interpolation, and morphological operations. One of the first examples of a digital workflow for splint design and additive manufacturing was presented by Salmi et al. \cite{Salmi2013}, who used \textit{ VisCam RP} software along with 3D scanning and stereolithography. In their six-month evaluation, the splints were reported to maintain dimensional stability and clinical usability. Adolphs et al. \cite{Adolphs2014} introduced a semi-automated approach, reducing the technician’s workload while preserving design flexibility.

A modular software prototype was proposed by Nasef et al. \cite{Nasef2014}, using segmentation of digital scans, occlusal contact analysis, and Boolean operations to generate the splint. The system was reported to show promising accuracy when tested on clinical cases.
Meanwhile, Chen and colleagues \cite{Chen2016feb,Chen2016} developed \textit{TemDesigner} and \textit{EasySplint}, which allowed semi-automated construction of surgical splints. These tools integrated user-defined landmarks, spline-based modeling of the external surface, and Boolean subtraction of dental surfaces to generate a stable splint that locks onto the teeth without movement.

\subsection{Fully digital workflows}

The transition to fully digital processes was exemplified by Waldecker et al. \cite{Waldecker2019}, who described an automated method for Michigan-type splints. Using intraoral scans and a registration jig for jaw positioning, the splint was designed in \textit{Zirkonzahn Modeller} \cite{Zirconzehn} and reported minimal need for clinical adjustment. Chen et al. \cite{Gao2024} later introduced \textit{OrthoCalc}, a 3D tool for evaluating maxillary position in six degrees of freedom. Sun et al. \cite{Sun2024} developed a fully automated pipeline and reported clinically accurate results. Likewise, Monaco et al. \cite{Monaco2018} described a consistent digital protocol, which was reported to offer more reproducible outcomes than manual techniques.

\subsection{AI-based methods}

Recent advances have introduced artificial intelligence (AI) into the design process of occlusal splints. Many commercial systems advertise AI-based functionalities but often do not disclose how or where AI is applied, nor the nature of the underlying training data. This lack of transparency has been noted in the literature as raising concerns regarding responsibility, as essential therapeutic decisions may be delegated to opaque algorithms. 

Several software platforms now incorporate AI to streamline the workflow, for example, \textit{3Shape Automate} and \textit{3Shape Splint Studio}, \textit{SprintRay Cloud Design}, \textit{Medit Splints}, \textit{Exocad}. 
Solutions are offered that enable both full automation of the occlusal splint design process, as well as those that allow the user to manually adjust individual steps of the design. These tools illustrate a growing trend toward partial or full automation of appliance design, combining speed with reproducibility.

However, Kois and Revilla-León \cite{Kois2023} compared AI-generated splints with those created by experienced technicians, and reported that although AI enables rapid and standardized design, notable differences remain in areas requiring anatomical individualization.

Overall, recent AI-driven solutions highlight an important trend toward automation, reproducibility, and reduced operator dependency. While these approaches are still evolving, they demonstrate the potential of AI to transform occlusal splint design and related workflows, particularly for scaling analyses and ensuring more standardized outcomes.

\subsection{Commercial software platforms}

Various commercial solutions now support occlusal splint design, either as dedicated systems or as plug-ins within broader CAD/CAM platforms. Popular tools include \textit{3Shape, Exocad, DentalCAD, MeditSplints} and \textit {Zirkonzahn}. Comparative studies have assessed differences in workflow and user experience \cite{Blasi2023,Shopova2022,Venezia2019}. For example, \textit{3Shape} was reported to demonstrate high precision and speed in splint design \cite{Blasi2023}, while \textit{Zirkonzahn Modeller} \cite{Waldecker2019} integrated full jaw registration and milling capabilities. Broader reviews of surgical planning software also indicate frequent clinical adoption of these platforms \cite{TEL2023775, LO2023100615}.

Together, these developments reflect a broader trend towards automation, accuracy, and standardization in occlusal splint production, driven by the convergence of CAD/CAM, 3D imaging, and AI technologies.

It should be noted that direct benchmarking against these proprietary platforms was not performed, as their closed-source nature and licensing restrictions preclude reproducible comparisons; instead, their capabilities are discussed qualitatively in the literature review.

\section{Background and Limitations of Previous Work}\label{sec:background}

\begin{figure}\centering
\includegraphics[width=0.8\textwidth]{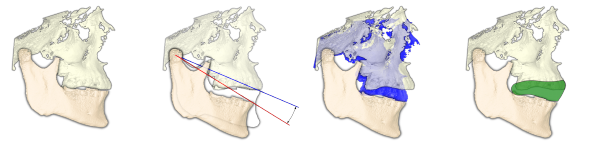}
\caption{From the left: 1.~Maxilla and mandible in maximum intercuspation. 2.~The therapeutic position of the mandible, 3.~Virtual movement of the occlusal surface into the therapeutic position. 4.~a splint built joining the  surface of maxilla teeth and moved occlusion surfaces.}\label{fig1234a}
\end{figure}

This article focuses on an occlusal positioning splint, building on and extending our earlier research on computational occlusal splint design and multimodal data integration \cite{Pojda2019,Tomaka2019}. In previous studies, the fundamental assumption was that both the mandible in a new (therapeutic) position (TP) and the splint itself should replicate the occlusal conditions of maximum intercuspation (MI) (Figure~\ref{fig1234a}). Throughout this paper, the term ''therapeutic position'' is used conventionally to denote the clinician-prescribed mandibular position, without implying therapeutic outcome. Triangular meshes were used to represent dental surfaces.
Standard mesh processing and rigid-body transformations (using $4\times4$ matrices and ICP alignment) were applied as described in \cite{botsch2010polygon,besl1992method,rusinkiewicz2001efficient}.

\begin{figure}\centering
\includegraphics[width=0.5\textwidth]{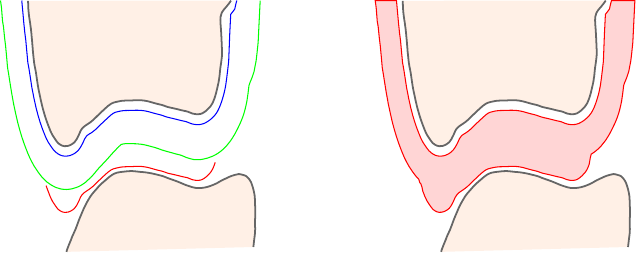}
\caption{Concept of the positioning splint: on the left: cross-sections of the upper and lower teeth shown in black, the occlusal surface in red, the inner (tooth-adapted) surface in blue, and the minimum thickness boundary (external surface) in green, on the right the resulting occlusal splint shown in red between the upper and lower teeth in black. }
 \label{fig:schemat}
\end{figure}

The approaches described in \cite{Pojda2019,Tomaka2019}  usually involved constructing the splint as a combination of the inner and outer surfaces. The inner surface was generated by extending the anatomical crowns of the maxillary teeth to obtain a minimum clearance, followed by a virtual impression. The outer surface is a combination of an external surface created using a similar method but corresponding to the minimum thickness boundary of the splint and the occlusal surface of the upper teeth, displaced by a specific rigid transformation (see Figure~\ref{fig:schemat}). The solid was closed with a cutting plane.

\begin{figure}[]
    \centering
    \includegraphics[width=0.7\linewidth]{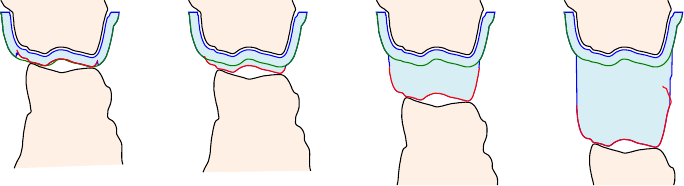}
    \caption{Occlusal splints for different displacements from the occlusal plane; the green line indicates the minimum splint thickness, and the red line marks the target therapeutic position (TP).}
    \label{fig:przeglad-app}
\end{figure}

Furthermore, the methods were validated mainly using simplified (translation only) transformations  (Figure~\ref{fig:przeglad-app}) and plaster models, without systematic assessment using real patient-derived 3D scans and dynamic jaw motion data. 

\begin{figure}\centering
\includegraphics[width=0.6\textwidth]{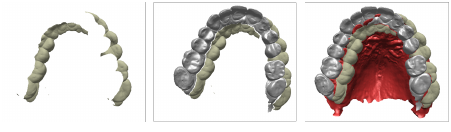}
\caption{From the left: 1.~External surface without surfaces shadowed by the occlusion in therapeutic position 2.~External and occlusal  surfaces together 3.~Cross-section of the splint with the upper teeth}
 \label{fig:szyna5}
\end{figure}

\begin{figure}\centering
\includegraphics[width=0.6\textwidth]{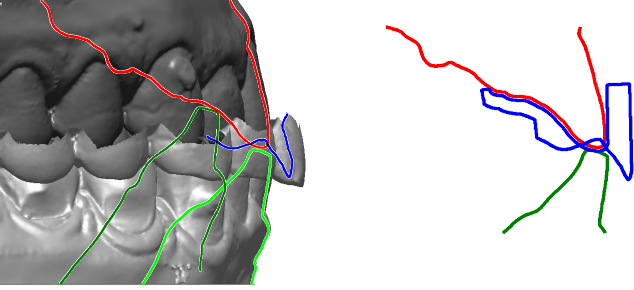}
\caption{Left: Profiles of the upper and lower teeth in maximal intercuspation (MI) and in the therapeutic position (TP); Right: the splint profile (blue) between the upper (red) and lower (green) teeth, with the cross-section taken through the upper teeth.}
 \label{fig:szyna5b}
\end{figure}

However, for real transformations, such workflows sometimes resulted in fragmented surfaces, self-intersections, or local collisions, making it difficult to create a functional, printable splint (Figure~\ref{fig:szyna5}). In many cases, it was not possible to fully preserve the original occlusal conditions in the therapeutic position, as shown by profile analyses (Figure~\ref{fig:szyna5b}).

\section{Advancing Methods for Designing Positioning Splints}\label{sec:model}

A key limitation of our previous work was that clinical feasibility was not always guaranteed, particularly when the prescribed transformation led to collisions, excessive proximity between surfaces, or other non-manufacturable geometries. To address these issues, the present study introduces a fully digital workflow that uses real patient data and a robust approach for resolving geometric conflicts and validating splint manufacturability and accuracy.

This work emphasizes interpreting the occlusal positioning splint as a tangible realization of a mathematical rigid-body transformation describing the change in maxilla--mandible relationship. Although absolute precision is not always necessary in practice, such splints are valuable whenever high accuracy in jaw positioning is required without the permanent fixation of surgical splints. This is especially relevant in multimodal image registration, where maintaining consistent anatomical relations across different imaging modalities (e.g., CBCT, MRI, and facial scans) is essential for reliable image integration and planning in research contexts.

To ensure that a positioning splint is both feasible and effective, we define the conditions for its construction, identify practical ways to acquire the intended transformation, analyze the range of realizable transformations, and present a general algorithm that allows for selective occlusal contacts through designated cutouts in the splint surface. We also verify the accuracy with which the physical splint replicates the prescribed mandibular repositioning.

\subsection{Preliminary assumptions}
One fundamental assumption in designing a positioning splint is the existence of a stable and repeatable occlusion. Maximum intercuspation (MI) is recognized as a natural, reliable jaw position, typically reproduced during involuntary actions such as swallowing \cite{ABDELHAKIM198212,Solow2015}. This stability is essential for establishing a predictable therapeutic relationship between the jaws.

The occlusal positioning splint is thus designed to reproduce a specific rigid-body transformation of the mandible, reflecting only the global jaw relationship rather than repositioning individual teeth. The splint should fit closely to the maxillary teeth to ensure secure placement, while allowing some freedom of mandibular movement. For this reason, it is not advisable to use negative impressions of both dental arches, as is common in surgical splints, which would lock the jaws and prevent physiological movement \cite{Chen2016}.

\subsection{Acquisition of the Therapeutic Transformation}

The therapeutic mandibular position is a clinician-defined, idealized spatial relationship between the maxilla and mandible, often selected to optimize function, comfort, or joint health, particularly in the management of temporomandibular disorders (TMD). In this work, the term ''therapeutic position'' is employed in a conventional sense, referring to a clinician-prescribed mandibular relation, without implying therapeutic effectiveness. Unlike habitual occlusion (maximum intercuspation, MI), which reflects the patient’s natural bite, the therapeutic position may coincide with centric relation (CR), but can also differ depending on individual clinical needs. It can be established by clinical evaluation (e.g., muscle palpation, joint auscultation) and transferred to the digital workflow using bite records, mandibular tracking, or by adjusting virtual models in CAD software.

\begin{figure}[t!]
	\centering
 	\includegraphics[width=0.65\textwidth]{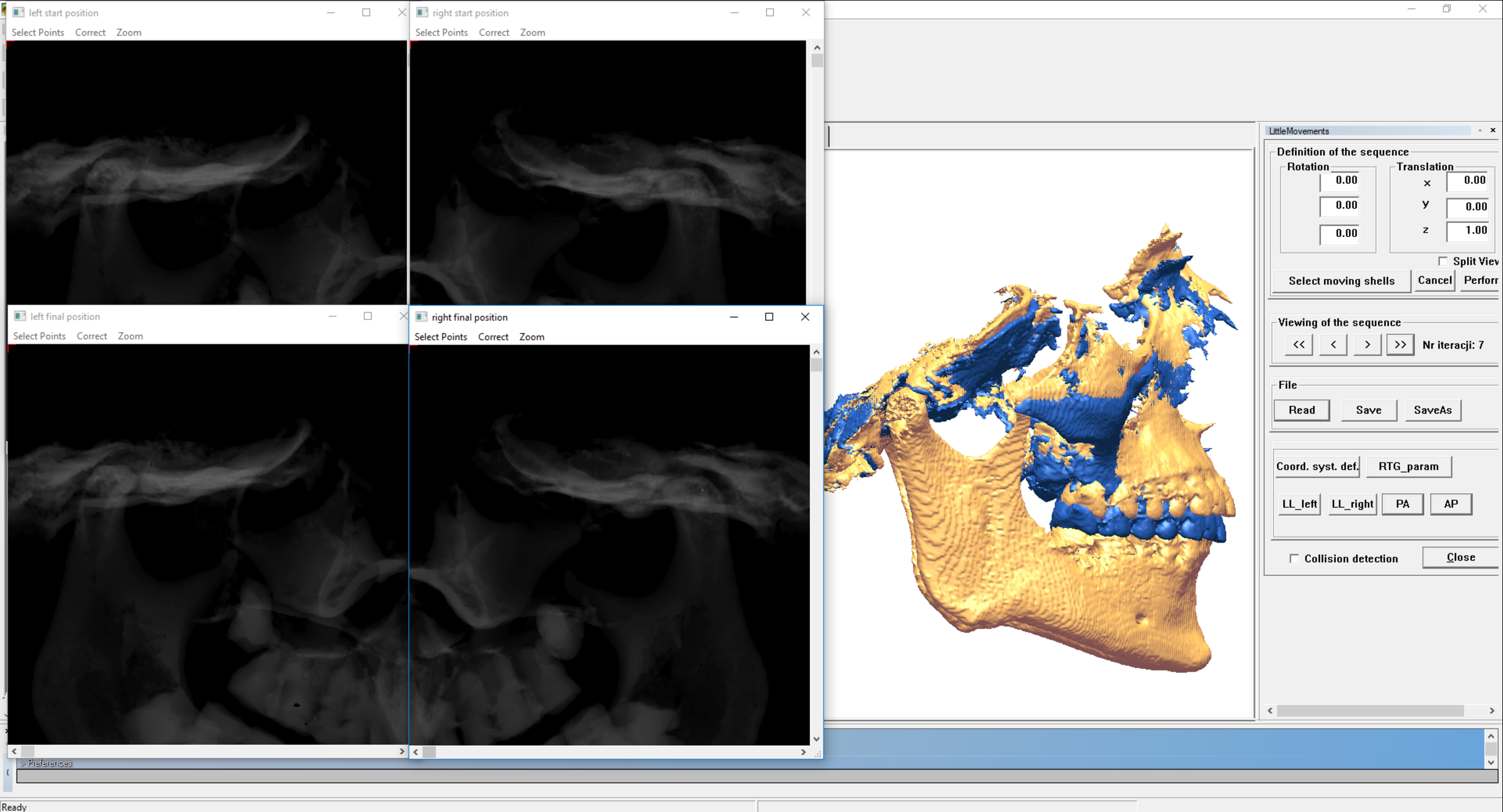}
	\caption{Editor of the mandibular movements; for each step, virtual X-rays of both condyles are computed.}
	\label{fig:littlemovements}
\end{figure}

In digital workflows, the therapeutic transformation ($T_{th}$), defined as the rigid-body movement that aligns MI to the therapeutic position (TP), can be obtained in several ways. One common approach is manual adjustment of virtual 3D models in CAD or dental CAD software, such as \textit{SICAT} \cite{Aslanidou2017} or \textit{EXOCAD}, where the clinician repositions the mandibular model to reflect the desired occlusion. The software then calculates the corresponding transformation matrix, sometimes with integrated visualization of TMJ dynamics to support clinical decisions (Figure~\ref{fig:littlemovements}).

\begin{figure}
    \centering
    \includegraphics[width=0.8\linewidth]{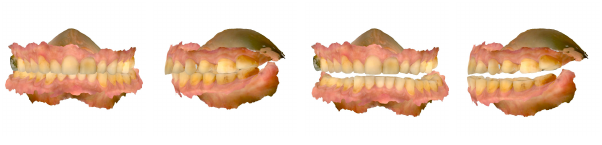}
\caption{Intraoral scans in two bites: maximal intercuspation (left) and a clinician-defined therapeutic position (right).}
\label{fig:intraoral}
\end{figure}

\begin{figure}[t!]
    \centering
    \includegraphics[width=0.75\linewidth]{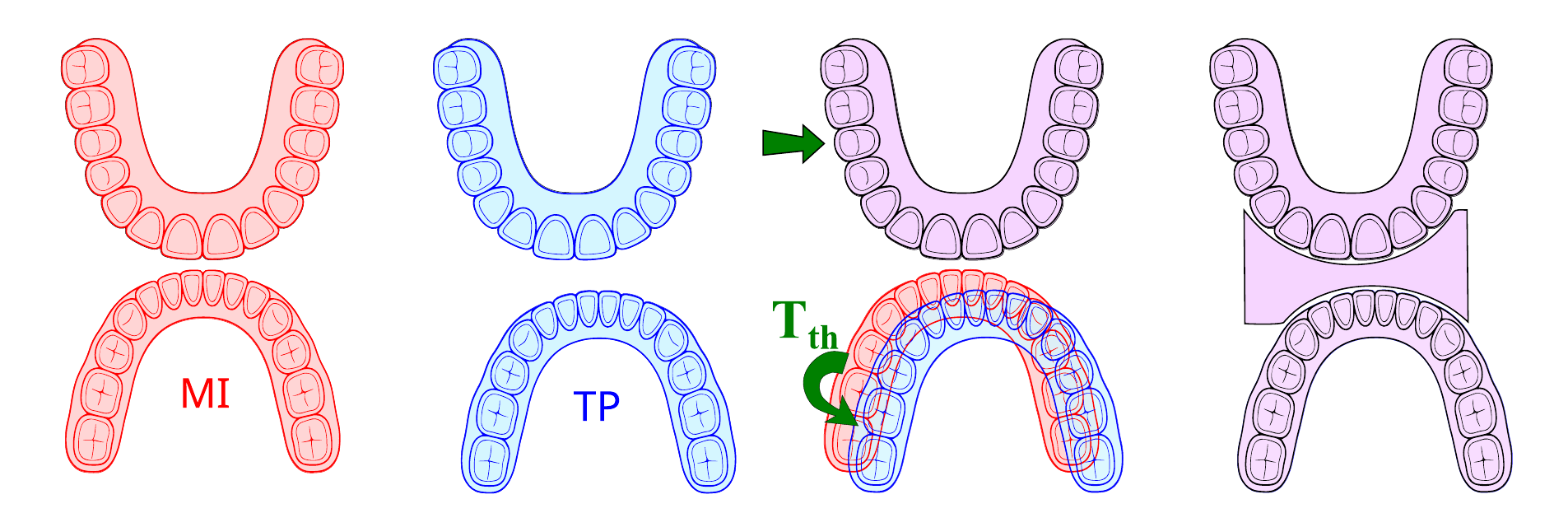}
    \caption{ From the left: diagrams of dental arches for maximum intercuspation (MI) and in the therapeutic position (TP), their registration with upper jaw alignment in order to determine the $T_{th}$ transform. Occlusal splint is a tangible physical realization of this transform, which preserves MI occlusion between the splint and the mandible and TP maxilla-mandibular relationship.}
    \label{fig:fig:szyna_mi2tp}
\end{figure}

Intraoral scanners offer a direct means to acquire both the 3D anatomy of the dental arches and their spatial relationships in different bite registrations (Figure~\ref{fig:intraoral}). By performing scans in MI and the therapeutic position, the required transformation can be computed by registering the mandibular models (Figure~\ref{fig:fig:szyna_mi2tp}) (using e.g., ICP) between reference coordinate systems. Details of this computation are provided in \ref{app:scanner}.

\begin{figure}[t!]
\centering
    \includegraphics[width=.8\linewidth]{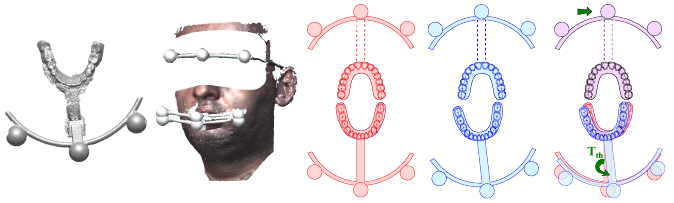}
    \caption{Maximal intercuspation and therapeutic mandibular position identified through 3D facial scans with a mandibular bow, serving as an interface between face and dental coordinate systems.}
    \label{fig:bow}
\end{figure}

Alternatively, mandibular tracking devices, such as a facial bow, can be used to record jaw movements, identifying MI and TP within the recorded sequence. By aligning the bow's position with intraoral scans, a transformation matrix relating the two positions can be determined (Figure~\ref{fig:bow}). The mathematical details of this procedure are given in \ref{app:tracker}.

\subsection{Constraints on the Therapeutic Transformation for Feasible Splint Realization}

Although therapeutic transformations obtained via intraoral scanners or mandibular tracking devices are generally limited by anatomical movement and registration accuracy, transformations freely defined in CAD software may produce physically unrealizable geometries. This risk increases when there are no restrictions on the range of mandibular displacement.

For a positioning splint to be constructed and used, several geometric conditions must be met. The maxillary and mandibular surfaces must not intersect after transformation, as any overlap would cause collisions during fabrication or use. The mandible should remain below the maxilla and within physiologically plausible distances, and mandibular rotation must stay within physiologically plausible limits to avoid unachievable joint positions.

Beyond these geometric rules, various anatomical and practical constraints may affect the feasibility and safety of the final design. Importantly, if only dental models are used for planning, certain critical risks may go undetected: for example, without CBCT-based bone registration, the condyle could penetrate the glenoid fossa after transformation, even when no occlusal collision is visible. Such a scenario may pose a potential risk if applied clinically, as a splint could drive the mandible toward a non-physiological position. This reinforces the rationale for preclinical use with anatomical imaging when exploring larger transformations. Therefore, whenever possible, splint design should rely on real jaw motion measurements and anatomical imaging of the TMJ. Table~\ref{tab:constraints} summarizes common causes of failure and their consequences.

\begin{table}[t!]
\centering
\caption{Constraints on the realizability of therapeutic mandibular transformation.}
\label{tab:constraints}\fontsize{9}{10}\selectfont
\begin{tabular}{@{}p{4cm}p{4cm}p{5cm}@{}}
\toprule
\makecell{\textbf{Constraint} \\ \textbf{/ Requirement}} & \makecell{\textbf{Cause of} \\ \textbf{Violation}} & \makecell{\textbf{Consequence} \\ \textbf{/ Risk}} \\
\midrule
Sufficient mandibular clearance in TP & Too small interocclusal distance (e.g. <1 mm) & Impossible to design a stable splint without collisions \\
\addlinespace
No intersection of upper and lower meshes after transformation & Excessive mandibular shift / rotation & Splint will penetrate anatomy; digital design fails \\
\addlinespace
Transform is physically achievable from MI & Extreme mandibular retraction or asymmetry & TMJ overload; splint may not fit or function properly \\
\addlinespace
Uniform surface contact between arches & Misaligned bite registrations or anatomical anomalies & Splint will be unstable or ineffective therapeutically \\
\addlinespace
No collision of mandibular and glenoid fossa bone structures & Use of dental models only, without CBCT or TMJ registration & Undetected TMJ penetration; high risk of patient injury \\
\bottomrule
\end{tabular}
\end{table}

\subsection{Method for Positioning Splint Design}

Having defined the anatomical and geometric constraints, we now present the improved workflow for constructing positioning splints. This method introduces a novel way for generating the outer surface, preserving occlusal contacts, and ensuring manufacturability.

\subsubsection{Segmentation and Reference Surfaces}

The design starts with 3D models of the dental arches. The relevant regions are identified by selecting the crown area of the maxillary teeth (above the undercuts), which serves as the basis for the inner and outer surfaces of the splint. The occlusal surface of the maxilla, after being transformed into the target therapeutic position \(\mathbf{T}_{th}\), is used to define the area of intended contact with the lower jaw (see Figure~\ref{fig:fig:szyna_mi2tp}). Segmentation is performed in a common reference system, with all data aligned accordingly.

\subsubsection{Inner and Outer Surface Construction}

\begin{figure}[t!]
    \centering
    \includegraphics{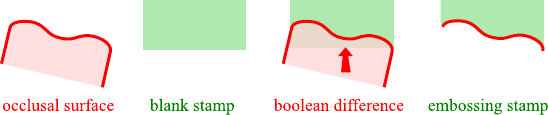}
    \caption{Creation of the digital embossing stamp from the transformed occlusal surface.}
    \label{fig:making-stamp-app}
\end{figure}

\begin{figure}[t!]
    \centering
    \includegraphics{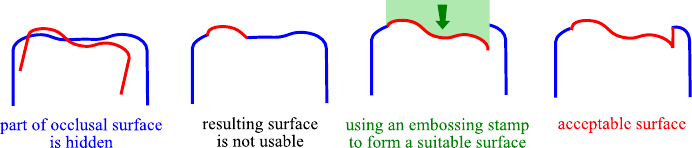}
    \caption{Imprinting the occlusal contact with the digital stamp.}
    \label{fig:using-stamp-app}
\end{figure}

\begin{figure}[t!]
    \centering
    \includegraphics{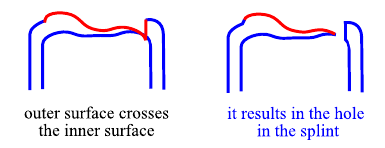}
    \caption{Illustrative example of an intersection between outer and inner surface resulting in an aperture (hole) in the splint.}
    \label{fig:hole-splint-app}
\end{figure}

\begin{figure}[t!]
    \centering
    \includegraphics[width=.8\linewidth]{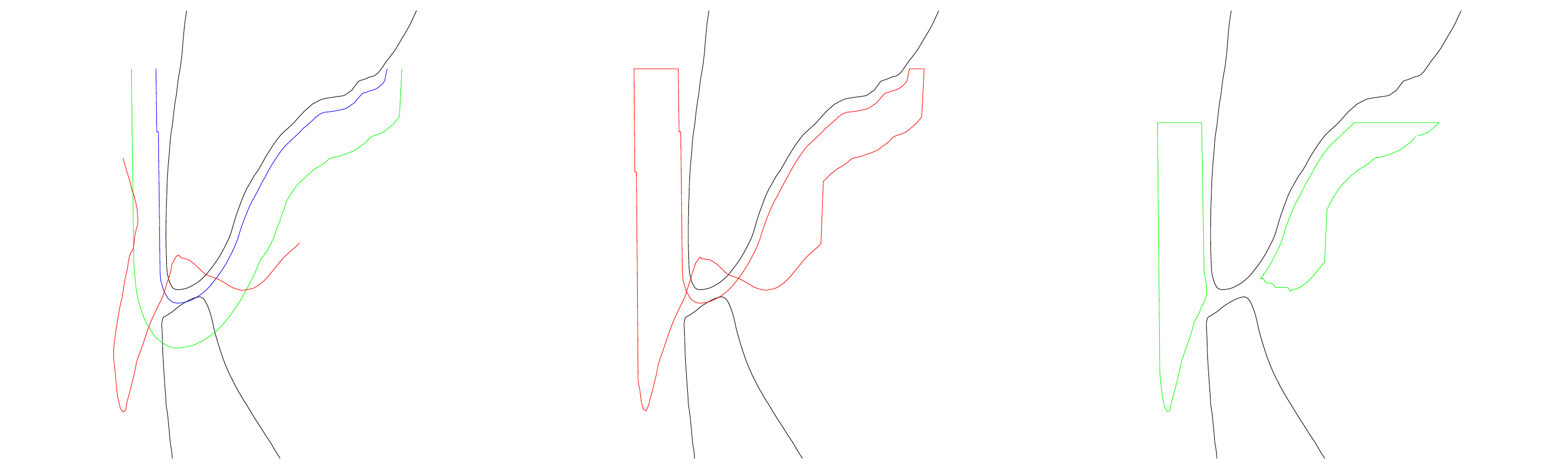}
    \caption{Left: profiles of the outer (green) and inner (blue) surfaces for a minimum-thickness splint on the maxilla; the target occlusal surface in the therapeutic position (TP) is shown in red. Center and right: profiles of fabricated splints without (center) and with (right) virtual stamp correction.}
    \label{fig:true-hole-app}
\end{figure}

Following our earlier approach, the inner splint surface is generated by dilating the coronal mesh to provide the necessary clearance and applying a virtual impression to restore surface regularity. The outer surface is constructed by further dilation, but this process alone may distort the occlusal contact area. As a key novelty of this work, we introduce a virtual embossing technique: a digital stamp is created from the occlusal surface in the therapeutic position (see Figure~\ref{fig:making-stamp-app}). The stamp is cleaned to remove artifacts from interdental spaces, then pressed into the outer shell, precisely restoring the target contact pattern (see Figure~\ref{fig:using-stamp-app}). To further avoid conflicts with the mandible, after embossing, an additional impression is made using the mandibular mesh in the therapeutic position as a stamp. When geometric conflicts occur (such as overlaps or intersections with the inner surface), the software automatically detects these regions and creates apertures to preserve both anatomical and manufacturing feasibility (Figures~\ref{fig:hole-splint-app}, \ref{fig:true-hole-app}).

\subsubsection{Finalization and Export}

The final splint is obtained by joining the inner and outer surfaces along the cutting plane. In regions where apertures have been created, the surfaces are connected to form additional boundary faces. The design is then validated, and any remaining conflicts are resolved as described above. The completed mesh can be exported for additive manufacturing.

\subsection{Practical Implementation: The \textit{Splint-maker} Plugin}

\begin{figure}
\centering
\includegraphics[width=.65\linewidth]{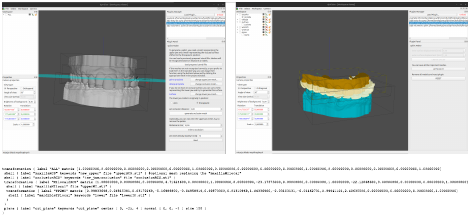}
\caption{Interface of the \textit{Splint-maker} plugin in \textit{dpVision}. Left: loaded input data (maxilla, mandible, occlusal surface) and plugin panel for mesh selection and parameter configuration. Right: generated splint placed on the upper jaw, with mandible in the therapeutic position. Bottom: sample input data description file}
\label{fig:splint-maker}
\end{figure}

\textit{Splint-maker} is a dedicated, open-source plugin for \textit{dpVision} software \cite{POJDA2025102093}, enabling semi-automatic construction of therapeutic occlusal splints based on the algorithms described here.
The process operates on 3D triangular meshes of the maxilla, the mandible (in either the initial or therapeutic position), and a trimmed occlusal surface (see Figure~\ref{fig:splint-maker}). A transformation matrix specifying the required mandibular displacement to the alternate position is also provided.

Input data can be loaded as a \texttt{.dpw} file, containing all required meshes and spatial metadata, enabling automatic mesh identification and role assignment (see Figure~\ref{fig:splint-maker}). Alternatively, the user can manually import and position models and configure options within the plugin.

If the occlusal surface is missing, the plugin can generate it automatically by analyzing the distance between the splint and mandible in the initial position, using a specified clearance to ensure functional contact in the therapeutic position.

Construction parameters, including mesh resolution (default 0.1~mm), cutting-plane depth, minimum wall thickness, and occlusal clearance, are configurable to meet clinical and manufacturing requirements. Splint generation starts with a single click and typically completes within a few minutes. The resulting model can be saved for analysis or fabrication.

In our implementation, the complete workflow (including data import and splint generation, but excluding 3D printing and scanning, which depend on external hardware) typically finishes within 2--3 minutes on standard desktop hardware (Intel i7 CPU, 32~GB RAM, no GPU acceleration). This confirms that the method is computationally feasible for practical use.

\section{Validation Experiments and Accuracy Analysis}\label{sec:results}

The aim of this experiment was to verify whether the designed and 3D-printed occlusal splint enables precise and repeatable reproduction of the planned therapeutic mandibular position relative to the maxilla. The validation was performed as a non-clinical demonstration, combining experimental fabrication and placement in a volunteer under laboratory conditions, followed by a multi-stage accuracy analysis covering different aspects of splint performance.

\subsection{Experimental Design and Data Acquisition}

\begin{figure}
    \centering
    \includegraphics[width=.9\linewidth]{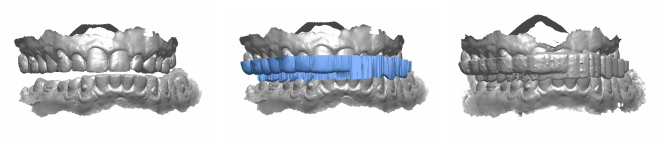}
    \caption{From the left:
1.~Intraoral scans showing the therapeutic relationship between the jaws. 
2.~Digital design of the occlusal splint for this jaw relation. 
3.~Intraoral scans of the volunteer with the 3D-printed occlusal splint in place (demonstration case, non-clinical).}\label{fig:scans_therap}
\end{figure}

The process began with 3D scanning of a volunteer’s dentition in maximal intercuspation (MI) and in the planned therapeutic position (TP).  The resulting triangular meshes of the maxilla and mandible, representing both positions, were registered into a common coordinate system using an initial alignment followed by refinement with the ICP algorithm.  An example of the input datasets and the jaw relationship in both positions is shown in Figure~\ref{fig:scans_therap}a.

Based on the registered datasets and the specified transformation, a digital model of the occlusal splint was designed and then manufactured using 3D printing (Figure~\ref{fig:scans_therap}b).  After the splint was produced, additional scans were performed of the dentition and the splint placed intraorally, including a follow-up scan with the teeth in occlusion on the splint (non-clinical demonstration) (Figure~\ref{fig:scans_therap}c).  All secondary scans were registered to the same MI reference frame, enabling direct comparison between the digital models and physical measurements. This also allowed assessment of how accurately the intended jaw relationship was reproduced.

\subsection{Accuracy Evaluation Procedures}

The accuracy assessment included several complementary analyses.  First, the geometric integrity of the splint was checked for unwanted collisions, self-intersections, and conflicts between the splint and tooth surfaces, both in the digital model before printing and in the physical model after fabrication.  This allowed us to exclude situations where the splint would be non-functional or would not meet clinical requirements.

\begin{figure}
    \centering
    \includegraphics[width=0.75\linewidth]{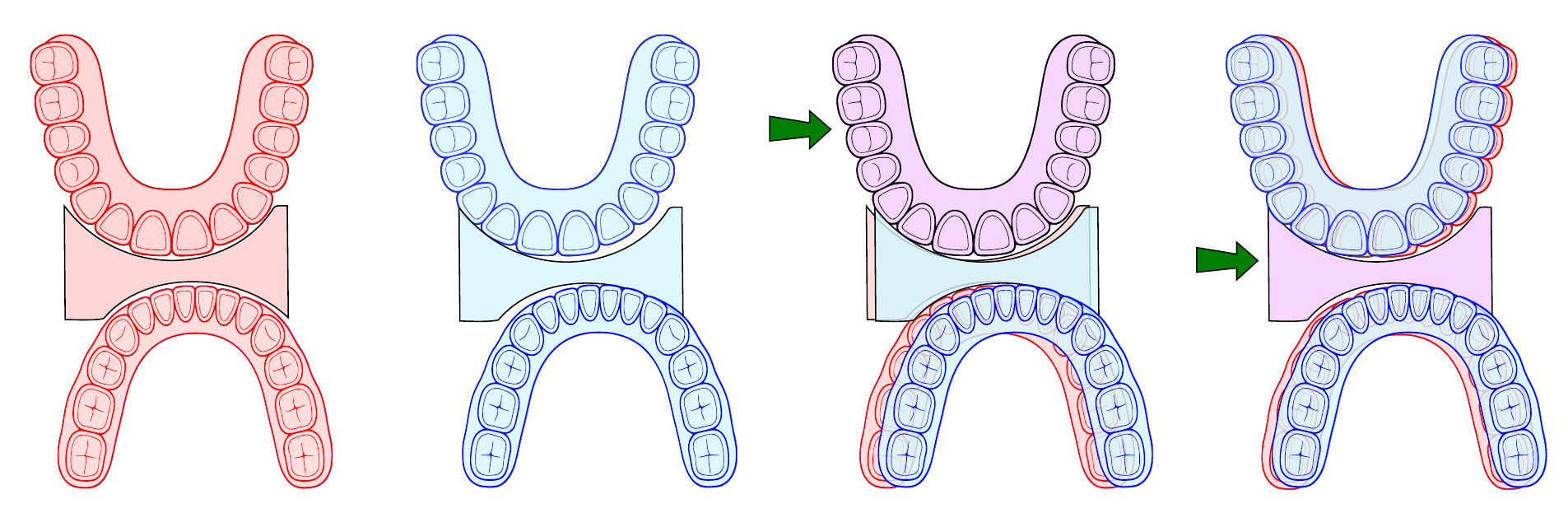}
    \caption{Accuracy assessment methodology. Scans acquired with the splint in place are registered to the reference coordinate system (RCS) to quantify the discrepancy between the prescribed and the resulting transformation. The splint scan is also registered to its digital model to estimate splint seating error on the maxilla and mandibular sliding on the splint.}
    \label{fig:szyna_blad}
\end{figure}

\begin{figure}
    \centering
    \includegraphics[width=.4\linewidth]{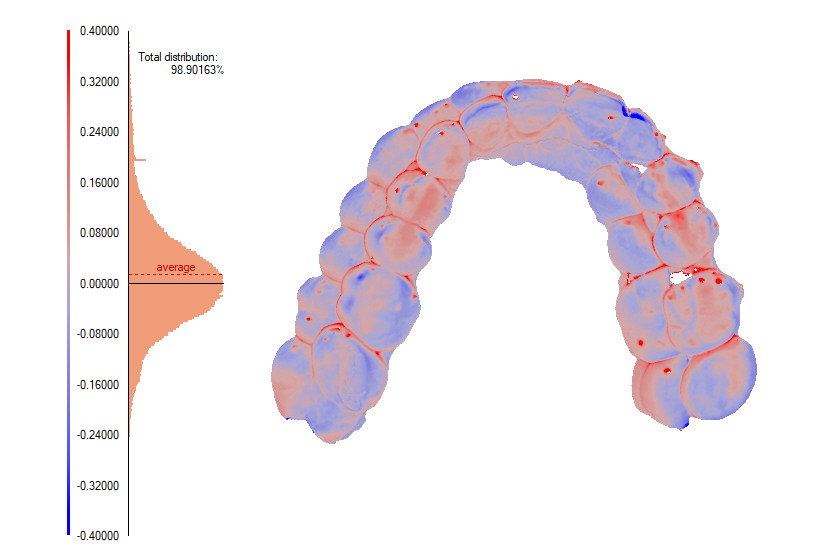}
    \caption{Visualisation of the accuracy of the 3D-printed splint compared to its digital design.
    }
    \label{fig:ocena01}
\end{figure}

The next stage involved a detailed analysis of the alignment between models obtained at different stages of the experiment (Figure~\ref{fig:szyna_blad}). This included both the accuracy of the 3D printing process and the precision in reproducing the intended jaw relationship. As a first step, the deviation of points from the scanned splint mesh was measured against its digital reference model, which allowed assessment of the overall accuracy of both fabrication and rescanning. The results of this analysis, shown as histograms of the signed fitting error and the corresponding standard deviations (STD), are presented in Figure~\ref{fig:ocena01}. In addition, the weighted mean of the signed error is reported to indicate any potential systematic bias.

Given the common modelling assumption that residuals are roughly normal (particularly within occlusal contact areas) we chose the standard deviation (STD) as a convenient summary statistic. Where this assumption is less plausible (e.g., peripheral mesh regions or isolated outliers, such as Splint~3), we discuss implications in Section~\ref{sec:discussion}. We do not assume strict normality of residuals; STD is used as a convenient summary under this common modelling assumption.

\begin{figure}
\centering
\includegraphics[width=.8\linewidth]{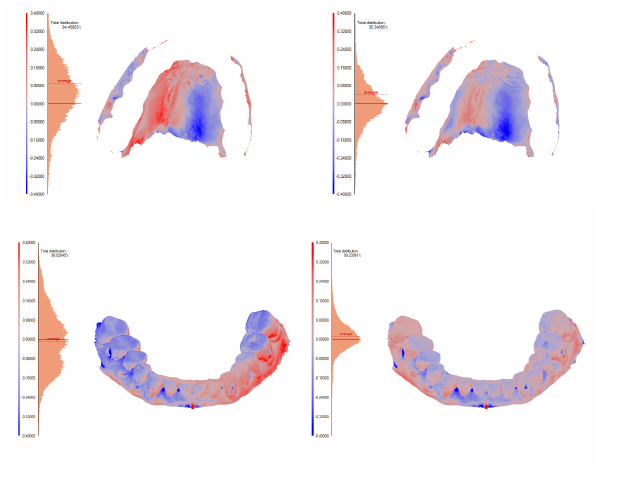}
\caption{Visualisation of: (a)~accuracy of the alignment between intraoral scans of the upper jaw and the corresponding reference model, based on fitting the scanned splint to its digital design;
(b)~accuracy after applying a corrective transformation modeling the disturbance of maxilla fitting in the splint;
(c)~accuracy of the alignment between intraoral scans of the lower jaw in therapeutic position and those obtained with the mandible positioned by the splint, before correction;
(d)~final mandibular positioning accuracy after applying the corrective transformation.}
\label{fig:ocena02}
\end{figure}

In the subsequent procedures, the fit of the maxilla to the splint was analyzed by isolating the palate area and computing a transformation that minimized the signed point-to-surface deviation relative to the original maxilla model.  This step estimates the clearance between the inner surface of the splint and the palate (Figures~\ref{fig:ocena02}a--b).  Next, the repeatability of mandible positioning on the splint was assessed by comparing the mandible scanned with the splint in place to its original therapeutic position.  Any sliding of the mandible on the splint surface was quantified by the STD of the signed fitting error, with the weighted mean serving to detect systematic offsets (Figures~\ref{fig:ocena02}c--d).

This multi-stage approach enabled us to evaluate fabrication accuracy and the reproducibility of the maxilla--mandible relationship after intraoral placement. It also allowed us to determine how clearances or shifts might affect the therapeutic outcome.

\begin{figure}
    \centering
    \includegraphics[width=0.65\textwidth]{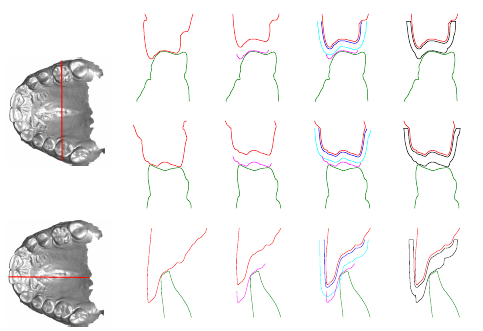}
    \caption{Analysis of profiles of the maxillary, inner, outer, and occlusal surfaces. From left to right: profile line; maxilla--mandible relationship; occlusal surface transformed into the therapeutic position (TP); inner and outer surfaces of the splint; resulting occlusal splint. Color coding: red, maxilla; green, mandible; dark and light blue, profiles of the minimum-thickness splint surfaces (inner and outer); pink, occlusal surface in TP; black, profile of the completed splint.}
    \label{fig:profiles_multi}
\end{figure}

The analysis was complemented by examining cross-sectional profiles in selected planes to inspect critical areas (such as the canines, molars, and palate) and to assess wall thickness and local gaps (Figure~\ref{fig:profiles_multi}). These profiles revealed local issues not captured by global statistics.

All analyses were documented both quantitatively (signed fitting error, AVG, and standard deviation, STD) and visually (distance maps, profiles, histograms), with the results summarized in tables and illustrated in figures.

\subsection{Results Summary}

\begin{figure}
   \centering
   \includegraphics[width=.8\linewidth]{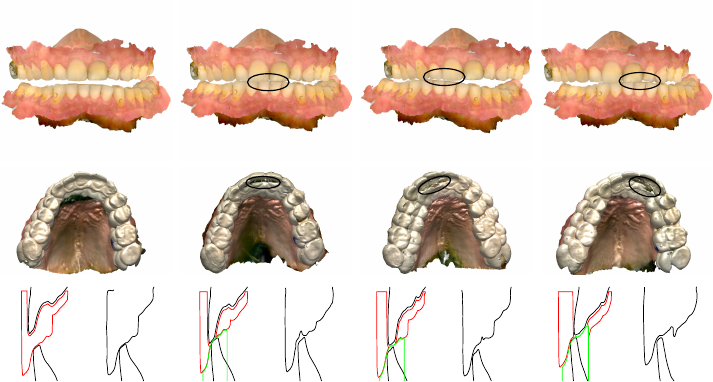}
   \caption{Different maxillomandibular relationships and the corresponding maxillary occlusal splints are assessed intraorally. The first splint (left) was generated for a no-tooth-contact relationship using a basic design algorithm. The remaining three splints, which incorporate openings to enable tooth contact, were created using the stamp-modified method. Verification is performed by profile analysis within the contact regions. The upper and lower arches (from both the first and second intraoral scans) are shown in black; the splint in red; and the stamp in green.}\label{fig:relationships}
\end{figure}

Eight positioning splints corresponding to different maxillomandibular relationships were fabricated for a single volunteer. Figure~\ref{fig:relationships} presents exemplary splints for non-tooth contact maxillomandibular relationships, which can be designed using the basic method, and for those with incisor contacts, which require the modified method.

For each splint, follow-up intraoral scans were performed, and data were analyzed using six complementary procedures: reproduction of the splint, reproduction of the maxilla and mandible, maxillary clearance, mandibular sliding, and final mandibular transformation.
Detailed per-splint results, including complete numerical tables and error-bar plots for each procedure, are provided in Appendix~B.  These results show that Splint~3 exhibited markedly higher dispersion in the stages of maxillary clearance and final mandibular transformation, which increased the pooled STD in these two procedures; therefore, Splint~3 is considered an outlier.

\begin{figure}[h!]
    \centering
    \includegraphics[width=0.75\linewidth]{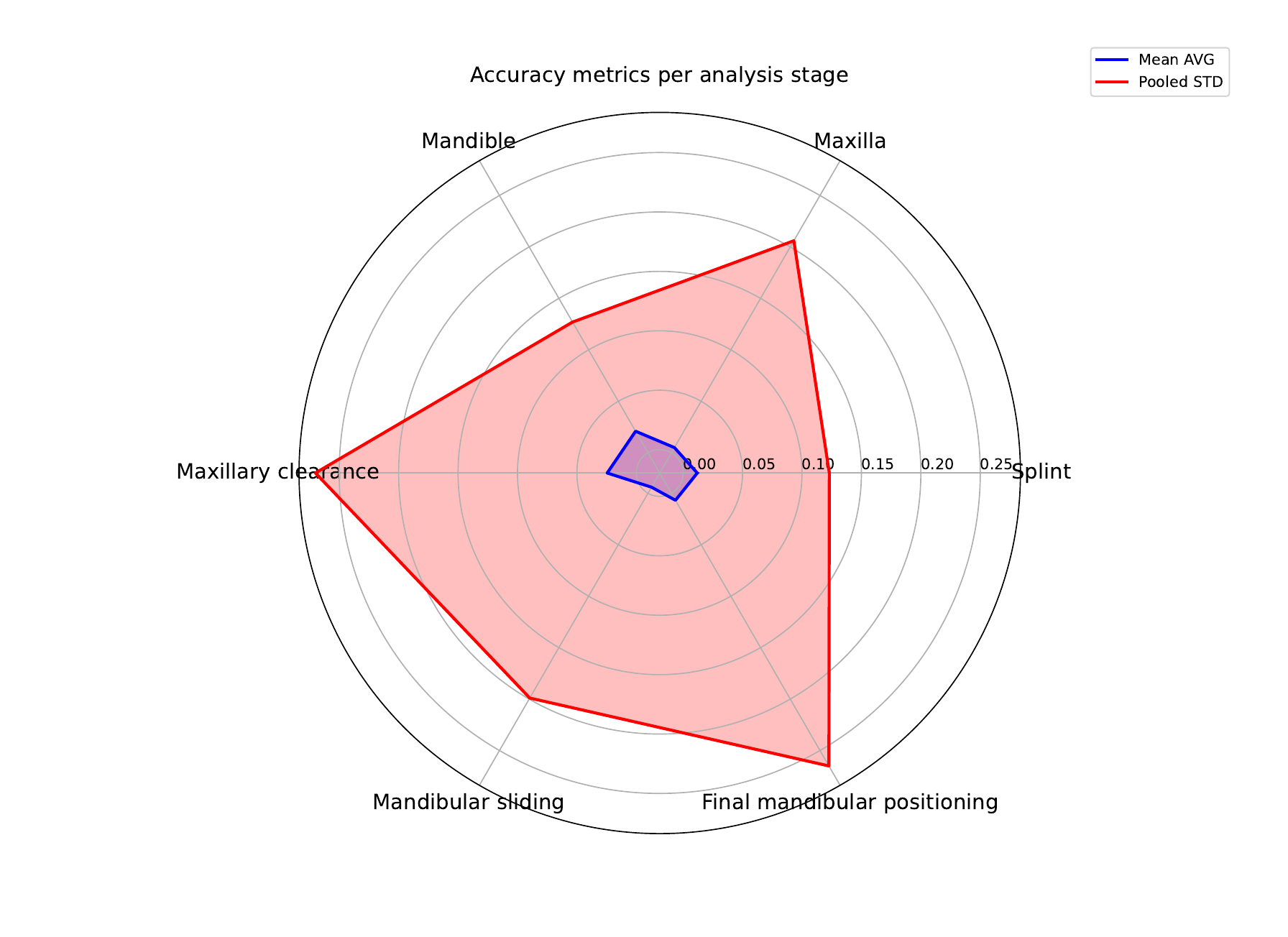}
    \caption{Radar plot showing the weighted mean of the signed fitting error (AVG) and pooled standard deviation (STD) for each analysis stage. This plot summarizes overall accuracy and precision; see Appendix~B for detailed per-splint results.}
    \label{fig:radar_error}
\end{figure}

Table~\ref{tab:summary_updated} summarizes the weighted mean AVGs and pooled STDs across all procedures.  For all analyses except maxillary clearance and final transformation, the pooled STDs remain below $0.21$~mm, demonstrating high precision.  All weighted mean AVGs are close to zero ($<0.03$~mm), confirming the absence of systematic bias.  A radar plot (Figure~\ref{fig:radar_error}) provides a visual comparison of these metrics across stages, and the combined error-bar plot in Appendix~B (Figure~B.22) illustrates the distribution of errors for each splint and procedure.

\begin{table}[ht]
\centering
\caption{Summary of updated accuracy metrics across all evaluation stages. For each stage, the table reports the weighted mean of the signed error (AVG) and the pooled standard deviation (STD) across all points and splints. These values correspond to the updated analysis using signed fitting errors.}
\label{tab:summary_updated}
\begin{tabular}{l|cc}
\toprule
\textbf{Analysis stage} & \makecell{Weighted mean AVG\\$[mm]$} & \makecell{Pooled STD\\$[mm]$} \\
\midrule
Splint vs.~model & 0.0119 & 0.123 \\
Maxilla vs.~model & 0.0050 & 0.206 \\
Mandible vs.~model & 0.0209 & 0.127 \\
\midrule
Maxillary clearance & 0.0245 & 0.270 \\
Mandibular sliding & $-0.0059$ & 0.199 \\
\midrule
Mandibular transformation & 0.0068 & 0.265 \\
\bottomrule
\end{tabular}
\end{table}

\paragraph{Implications}  The pooled standard deviations in Table~\ref{tab:summary_updated} indicate that the overall fitting precision remains within a sub-millimeter range consistent with typical CAD/CAM manufacturing and scanning precision. The slightly higher values observed for maxillary clearance and final transformation reflect contributions from soft tissues and patient-specific variability, amplified by Splint~3.  In contrast, the reproduction of the splint and mandible showed very low dispersion ($\leq 0.13$~mm).  These findings support the robustness of the proposed workflow and suggest that signed fitting errors can be effectively summarized using the pooled standard deviation as the primary metric of accuracy.

\section{Discussion and Future Directions}\label{sec:discussion}

The preceding sections have presented the theoretical basis, implementation, and experimental validation of a computer-aided method for designing occlusal positioning splints. In this section, we reflect on the broader implications of the proposed approach, highlighting its novel contributions, current limitations, and potential for clinical and technical development. We also outline possible future applications of positioning splints beyond their original therapeutic purpose.

\subsection{Novelty and Contribution}

This work builds on our earlier research on computational modeling of dental occlusion and multimodal data integration \cite{Pojda2019, Tomaka2019, TomakaG2019, Tomaka2007}. In previous studies, we established geometric operations (mesh dilation and virtual impression) and multimodal image registration methods to analyze jaw relationships and tooth contact surfaces. We also developed techniques for integrating CBCT, 3D facial scans, and jaw motion data, and validated these approaches on demonstration cases.

The main novelty of the present work lies in the introduction of an end-to-end, computer-aided design workflow that enables the fabrication of occlusal splints based on a user-defined therapeutic transformation matrix, specified by the clinician using a virtual patient model. In this approach, the splint is treated as a physical instantiation of a rigid-body transformation. It enforces the prescribed interarch position and acts as an operator that implements the desired occlusal relationship.

We further extend our method by systematically investigating the geometric feasibility of such splints. Specifically, we identify and analyze cases in which the target jaw position cannot be realized due to collisions, excessive proximity between the occlusal surface and the inner or outer surfaces of the splint, or other manufacturing limitations. To address these situations, we introduce a profile-based analysis to detect geometric conflicts and a virtual embossing (''stamp'') technique to resolve them. This ensures the splint’s manufacturability and functional performance.

Another important contribution is the integration of real patient data for splint planning. Unlike previous studies, where we applied transformations to synthetic models or stone casts, our workflow enables the therapeutic transformation to be directly extracted from facial scans and intraoral scans obtained in various jaw positions, including dynamic mandibular movements.

Finally, we introduce a comprehensive methodology for assessing splint performance and accuracy. This includes separate quantitative evaluation of maxilla-to-splint fit, mandibular positioning repeatability, and the overall error in reproducing the intended transformation. The use of surface deviation maps, profile analyses, and targeted metrics enables objective validation of both the design and demonstration application stages.

In contrast to our earlier work, which focused primarily on geometric and registration aspects, this study offers a fully documented and reproducible process for preclinical validation of splint design. The proposed approach thus serves as a transparent and extensible foundation for future validation studies and potential adoption in clinical workflows.

This deliberate choice ensures that the proposed workflow remains fully transparent and reproducible, providing a foundation onto which future AI modules can be safely integrated.

\subsection{Limitations of the Method}

The accuracy of the method depends on the precision of digital procedures (such as registration and modeling) and on technical limitations, including printing tolerances and scanner resolution. Furthermore, the evaluation based on a single-patient study requires validation on a larger cohort to account for variability in mandibular structure.

In real-world applications, additional factors may influence the method's repeatability and accuracy. These include the elasticity of soft tissues, variations in patient positioning during scanning, and possible manual errors during the placement or adjustment of the splint. Inaccuracies in splint seating (such as gaps between the splint and the dentition) may result in incomplete reproduction of the intended therapeutic position. Such misfits not only affect the intended positioning outcome but can also distort the results of post-treatment scanning and, consequently, lead to erroneous conclusions in accuracy assessments.

Another challenge occurs in cases of incomplete dentition. Missing teeth may reduce the number of reference surfaces available for occlusal contact, leading to unpredictable behavior of the algorithm when determining the occlusal surface or aligning the mandible. This variability may compromise the stability of the splint and limit the generalizability of the method.

Future work should include clinical trials, evaluation of mechanical stability under functional loads, and assessment of the long-term reproducibility of mandibular positioning in real-world scenarios.

Across all splints, the largest deviations were consistently observed at the peripheral regions of the scanned meshes. These points affect the statistical distribution of residuals and, in some cases, produce deviations from normality, thereby inflating global error metrics. Splint~3 illustrates an extreme case of this effect, where additional inaccuracies in the acquisition and registration of the upper jaw mesh amplified the overall dispersion. Such findings suggest that more advanced filtering of peripheral points should be considered in future studies. Importantly, this also implies that the actual error within 
contact areas is lower than the global metrics reported here.

Moreover, the present validation was limited to eight splints fabricated for a single volunteer, which constrains the statistical robustness of the findings. Although the multi-stage accuracy analysis provides objective insights, larger-scale validation on clinical patient cohorts will be required to establish the broader generalizability of the method.

Finally, direct benchmarking against commercial tools such as EasySplint or 3Shape was not performed. These platforms are proprietary and closed-source, limiting reproducibility. Instead, our study provides an open and transparent workflow that can serve as a reference baseline for future comparative evaluations.

\subsection{Further Research Directions}

The occlusal splint constructed in this work should be considered a \textit{positioning splint}, meaning that it is intended to enable the mandible to move to any physiologically possible position that the patient is capable of achieving. This includes, importantly, positions in which occlusal contact between the maxillary and mandibular teeth occurs. The primary goal in splint design is, therefore, to cover the broadest possible range of mandibular positions while maintaining mechanical stability and patient comfort.

In the short term, our ongoing work focuses on improving the construction of lateral and external surfaces (particularly in thicker splints), and on refining the handling of geometric conflicts that may arise in complex therapeutic transformations. These aspects are already being addressed in the continued development of the open-source \textit{Splint-maker} plugin.

In the longer term, additional research directions include identifying the geometric boundaries for feasible splint realization, determining the critical hole size or unsupported region that compromises mechanical stability, and exploring the design of sparse-contact splints. Finally, large-scale clinical evaluation remains an important challenge for the future, to be pursued in collaboration with medical partners.

Although our present workflow does not implement AI modules, we acknowledge that future integration of machine learning could further enhance segmentation, error detection, and quality control, thereby complementing the reproducibility offered by our transparent pipeline.

\subsection{New Applications Enabled by Positioning Splints}

The use of a positioning splint creates new possibilities that go beyond its basic therapeutic function. By stabilizing the mandible in a defined spatial relationship to the maxilla, the splint can serve as a reference object in various clinical and diagnostic contexts. One important application is the quantification of the CR/MI discrepancy. A splint designed to compensate for this discrepancy allows the corresponding transformation to be determined and used as a quantitative parameter describing the extent of occlusal mismatch.

In the field of multimodal registration, the splint can act as a rigid-body operator that imposes a known transformation on the mandible. This enables consistent patient positioning across different imaging modalities (such as CBCT, MRI, or optical scans). Similarly, when acquiring motion data, the use of a splint allows registration of coordinate systems from different mandibular motion acquisition devices, as the start and end positions can be precisely defined and reproduced.

The splint also facilitates consistent imaging, as scans can be acquired in a predefined jaw position. This improves comparability across sessions or devices, particularly in longitudinal studies or follow-up diagnostics. Moreover, the splint can be reused during clinical sessions to reproduce the same mandibular position, thereby increasing the repeatability and accuracy of therapeutic procedures.

Finally, the splint can be used as a comparative tool to evaluate different therapeutic strategies. By designing multiple splints that vary in their shape or in the jaw position they impose, researchers could in principle compare different splint designs or mandibular positions in terms of factors such as joint load, reproducibility, or potential comfort. Such comparative evaluation remains a topic for future clinical studies, and the present work provides a methodological foundation for these investigations.

\section{Conclusion}\label{sec:conclusions}

This study presents a comprehensive, computer-aided method for designing occlusal positioning splints that preserve or induce a therapeutic mandibular position defined by the clinician (used here in a conventional, geometric sense, without implication of therapeutic outcome). The splint is treated as a physical realization of a rigid-body transformation that accurately reproduces the prescribed change in the maxilla--mandible relationship.

The proposed approach integrates data from various imaging modalities (intraoral scans, CBCT, 3D facial scans) within a unified coordinate system, enabling precise reconstruction of the patient's anatomical structures and functional relationships. The method accounts for geometric constraints essential for the feasibility of splint design, including minimum printable thickness, accurate occlusal contact reproduction, and anatomical limitations.

In particular, we have introduced a mechanism for virtual embossing of the occlusal surface to accommodate therapeutic displacements that would otherwise intersect the external surface of the splint. This approach allows for high-precision realization of the therapeutic occlusion, even in complex cases. Additionally, we demonstrated multiple methods for acquiring the transformation matrix, including the use of intraoral scanners and mandibular tracking devices, and evaluated the accuracy of the resulting splints using digital and intraoral assessment techniques.

The described methodology enables reproducible splint design with relevance for clinical workflows in the future and offers a foundation for further applications, including multimodal image registration, quantification of CR/MI discrepancies, and comparative studies of different jaw repositioning strategies.

The presented approach opens several promising research directions aimed at extending the design and potential future clinical translation of positioning splints. Future work will focus on improving splint geometry, stability, and adaptability, as well as exploring their diagnostic value in multimodal image registration and functional evaluation of the stomatognathic system.

\section*{Ethical and Regulatory Note}
This work is a methodological, preclinical proof-of-concept. All scans were acquired in a non-clinical demonstration setting from a single healthy volunteer with non-therapeutic intent. No diagnostic or therapeutic decisions were made based on the presented methods, and no medical device claims are made. The procedures involved only optical scanning and 3D printing tasks typical for in-lab prototyping.

\section*{Code Availability}
The source code of the \textit{Splint-maker} plugin is openly available at \url{https://github.com/iitis/splint-maker}. 
The \textit{dpVision} platform, which provides the underlying framework, is also released as open-source software at \url{https://github.com/pojdulos/dpVision}.

\paragraph{Intended use and disclaimer}
The \textit{Splint-maker} plugin and the \textit{dpVision} platform are provided for research use only. They are not intended for diagnosis, treatment, or direct clinical decision-making. Any prospective clinical application would require appropriate validation, regulatory assessment, and oversight.

\section*{Acknowledgements}

The authors would like to thank Krzysztof Domino, PhD, for his valuable advice and motivation, which supported the creation of this article.


\appendix

\section{Estimation of Therapeutic Transformation}
\subsection{Intraoral Scanner-Based Estimation}
\label{app:scanner}

This appendix presents the mathematical derivation of the transformation matrix based on intraoral scan registrations.

Let us assume:
\begin{itemize}
    \item $\mathbf{U}_0$, $\mathbf{L}_0$ -- point clouds of the upper and lower dental arches acquired in the MI position, in the reference coordinate system (RCS),
    \item $\mathbf{U}_1$, $\mathbf{L}_1$ -- point clouds of the upper and lower arches in the therapeutic position (TP), in the measurement coordinate system (MCS),
    \item $\mathbf{T}$ -- the transformation from MCS to RCS,
    \item $\mathbf{L}_\text{th}$ -- the transformed lower arch in the TP position, expressed in RCS,
    \item $\mathbf{T}_\text{th}$ -- the final transformation matrix describing the mandibular shift from MI to TP in the RCS.
\end{itemize}

The estimation proceeds as follows:

\begin{enumerate}
    \item First, align the upper arches $\mathbf{U}_0$ and $\mathbf{U}_1$ to find the transformation $\mathbf{T}$ from MCS to RCS:
    \begin{equation}
        \mathbf{T} = \arg\min_{\mathbf{T}} \| \mathbf{U}_0 - \mathbf{T} \cdot \mathbf{U}_1 \|^2
    \end{equation}

    \item Apply $\mathbf{T}$ to the lower arch $\mathbf{L}_1$ in TP to bring it into the RCS:
    \begin{equation}
        \mathbf{L}_\text{th} = \mathbf{T} \cdot \mathbf{L}_1
    \end{equation}

    \item Estimate the therapeutic transformation $\mathbf{T}_\text{th}$ as the alignment from $\mathbf{L}_0$ to $\mathbf{L}_\text{th}$:
    \begin{equation}
        \mathbf{T}_\text{th} = \arg\min_{\mathbf{T}_\text{th}} \| \mathbf{L}_\text{th} - \mathbf{T}_\text{th} \cdot \mathbf{L}_0 \|^2
    \end{equation}
\end{enumerate}

All optimizations can be implemented using the Iterative Closest Point (ICP) algorithm, under the assumption of rigid-body transformations (rotation + translation). 

This method allows estimation of the therapeutic transformation based on two intraoral scans, without relying on external hardware.

\subsection{Mandibular Tracking-Based Estimation}
\label{app:tracker}

This section outlines the procedure for determining the therapeutic mandibular transformation using a facial bow and mandibular motion tracking.

Assume the following:
\begin{itemize}
    \item $\mathbf{B}_0, \mathbf{B}_1$ -- point clouds of the mandibular bow in the MI and TP positions, recorded in the measurement coordinate system (MCS),
    \item $\mathbf{T}_B$ -- transformation matrix representing the motion of the bow from MI to TP in MCS,
    \item $\mathbf{T}_F, \mathbf{T}_D$ -- transformations aligning the coordinate systems of the face scanner (FS) and dental model (DM) to the bow and teeth,
    \item $\mathbf{T}$ -- global transformation from FS to DM coordinate system.
\end{itemize}

The estimation proceeds as follows:

\begin{enumerate}
    \item Compute the transformation of the mandibular bow in MCS:
    \begin{equation}
        \mathbf{T}_B = \arg\min_{\mathbf{T}_B} \| \mathbf{B}_1 - \mathbf{T}_B \cdot \mathbf{B}_0 \|^2
    \end{equation}

    \item Compute the transformation between coordinate systems:
    \begin{equation}
        \mathbf{T} = \mathbf{T}_{D}^{-1} \cdot \mathbf{T}_F
    \end{equation}

    \item Transform the mandibular model from MI in DM space to FS space:
    \begin{equation}
        \mathbf{L}_0^\text{F} = \mathbf{T}^{-1} \cdot \mathbf{L}_0
    \end{equation}

    \item Apply the bow transformation in FS space:
    \begin{equation}
        \mathbf{L}_1^\text{F} = \mathbf{T}_{B} \cdot \mathbf{L}_0^\text{F}
    \end{equation}

    \item Transfer the transformed model back to DM coordinate system:
    \begin{equation}
        \mathbf{L}_1 = \mathbf{T} \cdot \mathbf{L}_1^\text{F}
    \end{equation}

    \item Derive the final therapeutic transformation matrix in DM space:
    \begin{equation}
        \mathbf{T}_\text{th} = \mathbf{T} \cdot \mathbf{T}_{B} \cdot \mathbf{T}^{-1}
    \end{equation}
\end{enumerate}

This procedure assumes that the coordinate systems of the tracking device and the intraoral scan can be related through proper calibration (i.e., rigid-body registration of bow surfaces and tooth scans). Once such registration is achieved, the bow acts as a physical linking element between the two domains.

\section{Detailed per-splint accuracy results}\label{app:results}

\begin{figure}[h!]
    \centering
    \includegraphics[width=\linewidth]{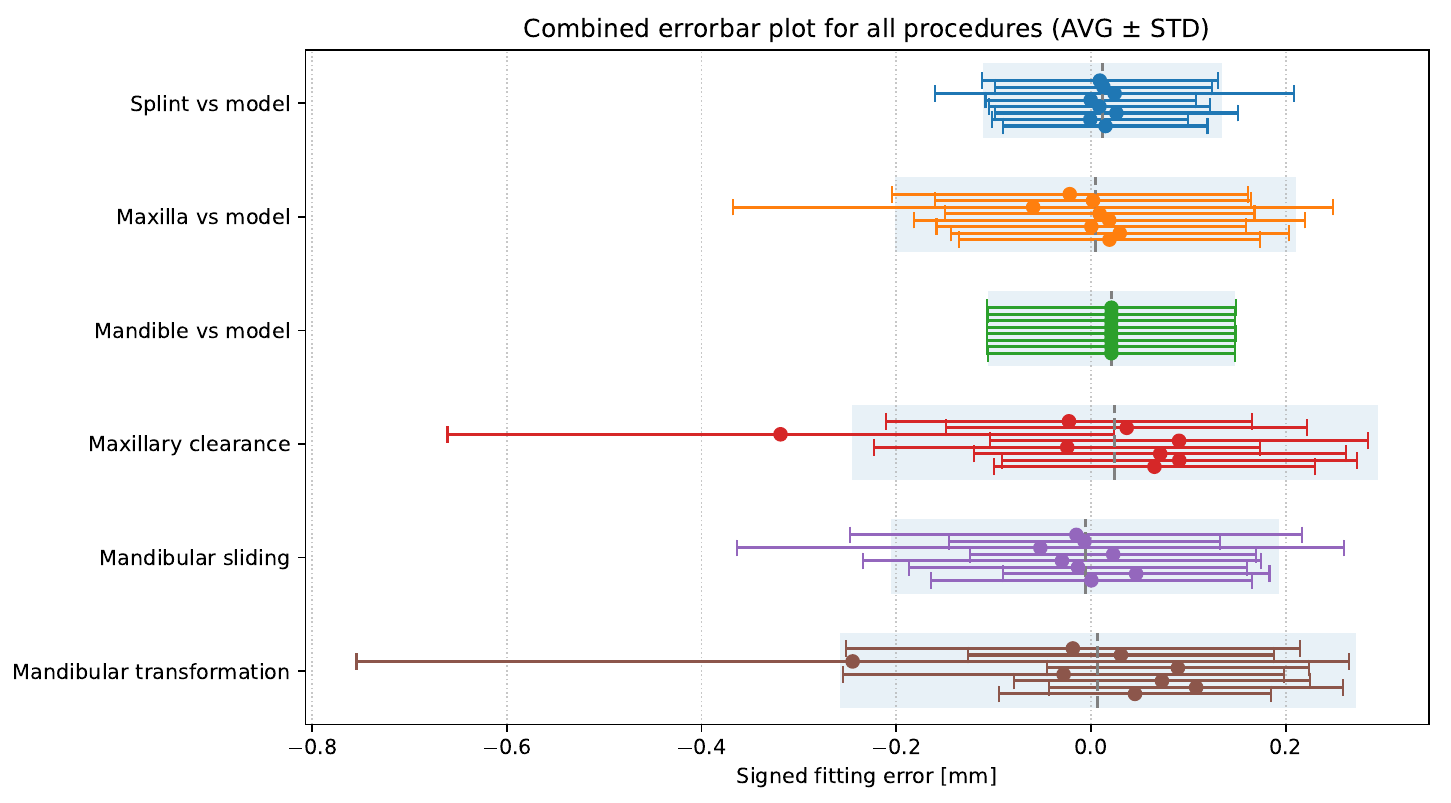}
    \caption{Combined errorbar plot for all six evaluation procedures. Each point shows the mean signed error (AVG) for one splint, with horizontal bars representing the standard deviation (STD). The shaded area marks $\pm 1$ pooled STD around the weighted mean AVG for each procedure, illustrating overall dispersion. This figure complements the radar plot in the main text by providing per-splint detail.}
    \label{fig:zbiorczy}
\end{figure}

Detailed results for all evaluation stages and all splints are presented in the following tables.
The graphical summary of all six procedures is provided in Figure~\ref{fig:zbiorczy}.

\begin{table}[ht]
\centering
\caption{Comparison of the scanned splint with its digital model. $N$: number of points; AVG: mean signed error [mm]; STD: standard deviation [mm]. STD is the main measure of fit accuracy; the weighted mean AVG (close to zero) indicates no systematic bias. Splint 3 presents as an outlier (higher STD).}
\label{tab:appendix_splint}
\begin{tabular}{c|c|c|c}
\toprule
\textbf{Splint} & $N$ & AVG [mm] & STD [mm] \\
\midrule
9 & 206456 & 0.0149 & 0.1048 \\
7 & 196232 & -0.0007 & 0.1007 \\
6 & 197783 & 0.0261 & 0.1246 \\
5 & 220863 & 0.0087 & 0.1135 \\
4 & 171191 & -0.0003 & 0.1080 \\
3 & 179972 & 0.0243 & 0.1842 \\
8 & 193355 & 0.0128 & 0.1113 \\
1 & 174612 & 0.0091 & 0.1211 \\
\midrule
Total $N$ & 1\,540\,464 & -- & -- \\
Weighted mean & -- & 0.0119 & -- \\
Pooled STD & -- & -- & 0.123 \\
\bottomrule
\end{tabular}
\end{table}

\begin{table}[ht]
\centering
\caption{Comparison of the scanned maxilla with its digital model. $N$: number of points; AVG: mean signed error [mm]; STD: standard deviation [mm]. STD is the main accuracy indicator; splint 3 stands out with the highest error.}
\label{tab:appendix_maxilla}
\begin{tabular}{c|c|c|c}
\toprule
\textbf{Splint} & $N$ & AVG [mm] & STD [mm] \\
\midrule
9 & 52062 & 0.0190 & 0.1544 \\
7 & 52179 & 0.0296 & 0.1737 \\
6 & 49651 & 0.0004 & 0.1591 \\
5 & 42597 & 0.0187 & 0.2008 \\
4 & 50569 & 0.0088 & 0.1592 \\
3 & 83456 & -0.0594 & 0.3084 \\
8 & 52587 & 0.0021 & 0.1619 \\
1 & 51778 & -0.0218 & 0.1830 \\
\midrule
Total $N$ & 434\,879 & -- & -- \\
Weighted mean & -- & 0.0050 & -- \\
Pooled STD & -- & -- & 0.206 \\
\bottomrule
\end{tabular}
\end{table}

\begin{table}[ht]
\centering
\caption{Comparison of the scanned mandible with its digital model. $N$: number of points; AVG: mean signed error [mm]; STD: standard deviation [mm].}
\label{tab:appendix_mandible}
\begin{tabular}{c|c|c|c}
\toprule
\textbf{Splint} & $N$ & AVG [mm] & STD [mm] \\
\midrule
9 & 158435 & 0.0210 & 0.1272 \\
7 & 158438 & 0.0210 & 0.1274 \\
6 & 158438 & 0.0209 & 0.1273 \\
5 & 158436 & 0.0208 & 0.1280 \\
4 & 158435 & 0.0209 & 0.1274 \\
3 & 158440 & 0.0209 & 0.1275 \\
8 & 158432 & 0.0210 & 0.1273 \\
1 & 158437 & 0.0210 & 0.1275 \\
\midrule
Total $N$ & 1\,267\,491 & -- & -- \\
Weighted mean & -- & 0.0209 & -- \\
Pooled STD & -- & -- & 0.127 \\
\bottomrule
\end{tabular}
\end{table}

\begin{table}[ht]
\centering
\caption{Maxillary clearance: fit of the splint on the maxilla (post-placement). $N$: number of points; AVG: mean signed error [mm]; STD: standard deviation [mm]; alpha: rotation error [$^\circ$]; t: translation error [mm]. Outlier: splint 3 (largest error).}
\label{tab:appendix_clearance}
\begin{tabular}{c|c|c|c|c|c}
\toprule
\textbf{Splint} & $N$ & AVG [mm] & STD [mm] & alpha [$^\circ$] & t [mm] \\
\midrule
9 & 52062 & 0.0652 & 0.1647 & 0.1651 & 0.1144 \\
7 & 52179 & 0.0907 & 0.1821 & 0.1172 & 0.1349 \\
6 & 49651 & 0.0711 & 0.1913 & 0.2448 & 0.2486 \\
5 & 42597 & -0.0244 & 0.1980 & 0.7186 & 0.4098 \\
4 & 50569 & 0.0905 & 0.1939 & 0.1003 & 0.2413 \\
3 & 83456 & -0.3189 & 0.3421 & 0.3620 & 0.6089 \\
8 & 52587 & 0.0367 & 0.1854 & 0.1571 & 0.1325 \\
1 & 51778 & -0.0226 & 0.1876 & 0.4929 & 0.2369 \\
\midrule
Total $N$ & 434\,879 & -- & -- & -- & -- \\
Weighted mean & -- & 0.0245 & -- & -- & -- \\
Pooled STD & -- & -- & 0.270 & -- & -- \\
\bottomrule
\end{tabular}
\end{table}

\begin{table}[ht]
\centering
\caption{Mandibular sliding: movement of the mandible on the splint. $N$: number of points; AVG: mean signed error [mm]; STD: standard deviation [mm]; alpha: rotation error [$^\circ$]; t: translation error [mm].}
\label{tab:appendix_sliding}
\begin{tabular}{c|c|c|c|c|c}
\toprule
\textbf{Splint} & $N$ & AVG [mm] & STD [mm] & alpha [$^\circ$] & t [mm] \\
\midrule
9 & 158435 & 0.0004 & 0.1648 & 0.3542 & 0.3033 \\
7 & 158438 & 0.0464 & 0.1370 & 0.0879 & 0.0908 \\
6 & 158438 & -0.0133 & 0.1740 & 0.2281 & 0.2330 \\
5 & 158436 & -0.0298 & 0.2044 & 0.5738 & 0.2123 \\
4 & 158435 & 0.0228 & 0.1468 & 0.2960 & 0.2060 \\
3 & 158440 & -0.0520 & 0.3119 & 0.4824 & 0.5132 \\
8 & 158432 & -0.0066 & 0.1394 & 0.2980 & 0.1698 \\
1 & 158437 & -0.0150 & 0.2322 & 0.3582 & 0.3054 \\
\midrule
Total $N$ & 1\,267\,491 & -- & -- & -- & -- \\
Weighted mean & -- & -0.0059 & -- & -- & -- \\
Pooled STD & -- & -- & 0.199 & -- & -- \\
\bottomrule
\end{tabular}
\end{table}

\begin{table}[ht]
\centering
\caption{Mandibular transformation: accuracy of final mandible-to-maxilla relationship. $N$: number of points; AVG: mean signed error [mm]; STD: standard deviation [mm]; alpha: rotation error [$^\circ$]; t: translation error [mm].}
\label{tab:appendix_transformation}
\begin{tabular}{c|c|c|c|c|c}
\toprule
\textbf{Splint} & $N$ & AVG [mm] & STD [mm] & alpha [$^\circ$] & t [mm] \\
\midrule
9 & 158435 & 0.0452 & 0.1401 & 0.4433 & 0.1979 \\
7 & 158438 & 0.1080 & 0.1509 & 0.1695 & 0.1466 \\
6 & 158438 & 0.0729 & 0.1521 & 0.2263 & 0.1067 \\
5 & 158436 & -0.0281 & 0.2267 & 0.4879 & 0.3897 \\
4 & 158435 & 0.0893 & 0.1344 & 0.2392 & 0.0600 \\
3 & 158440 & -0.2448 & 0.5096 & 0.5317 & 0.9831 \\
8 & 158432 & 0.0308 & 0.1568 & 0.3707 & 0.1665 \\
1 & 158437 & -0.0186 & 0.2330 & 0.1821 & 0.3122 \\
\midrule
Total $N$ & 1\,267\,491 & -- & -- & -- & -- \\
Weighted mean & -- & 0.0068 & -- & -- & -- \\
Pooled STD & -- & -- & 0.265 & -- & -- \\
\bottomrule
\end{tabular}
\end{table}

\clearpage
\end{document}